\documentclass[10pt,twocolumn,letterpaper]{article}

\usepackage[pagenumbers]{cvpr} %

\usepackage{times}
\usepackage{epsfig}
\usepackage{graphicx}
\usepackage{amsmath}
\usepackage{amssymb}
\usepackage{booktabs}
\usepackage{multirow}
\usepackage{xcolor}
\usepackage{bm}
\usepackage{multicol}
\usepackage[inline]{enumitem}
\usepackage{subfiles}

\usepackage{capt-of,etoolbox}
\makeatletter
\apptocmd\@maketitle{{\teaserfigure{}\par}}{}{}
\makeatother

\newcommand{\code}[1]{\texttt{#1}}
\renewcommand{\bar}[1]{\widetilde{#1}}
\renewcommand{\hat}[1]{\widehat{#1}}
\renewcommand{\paragraph}[1]{\noindent\textbf{#1}\enskip}
\newcommand{\metric}[1]{\textit{\textbf{#1}}}

\usepackage[pagebackref,breaklinks,colorlinks]{hyperref}

\usepackage[capitalize]{cleveref}
\crefname{section}{Sec.}{Secs.}
\Crefname{section}{Section}{Sections}
\Crefname{table}{Table}{Tables}
\crefname{table}{Tab.}{Tabs.}

\def\cvprPaperID{1802} %
\def\confName{CVPR}
\def\confYear{2022}

\newcommand{\teaserfigure}{
    \centering
    \vspace{-2.5em}
    \includegraphics[width=0.94\linewidth, trim=150 50 0 0, clip]{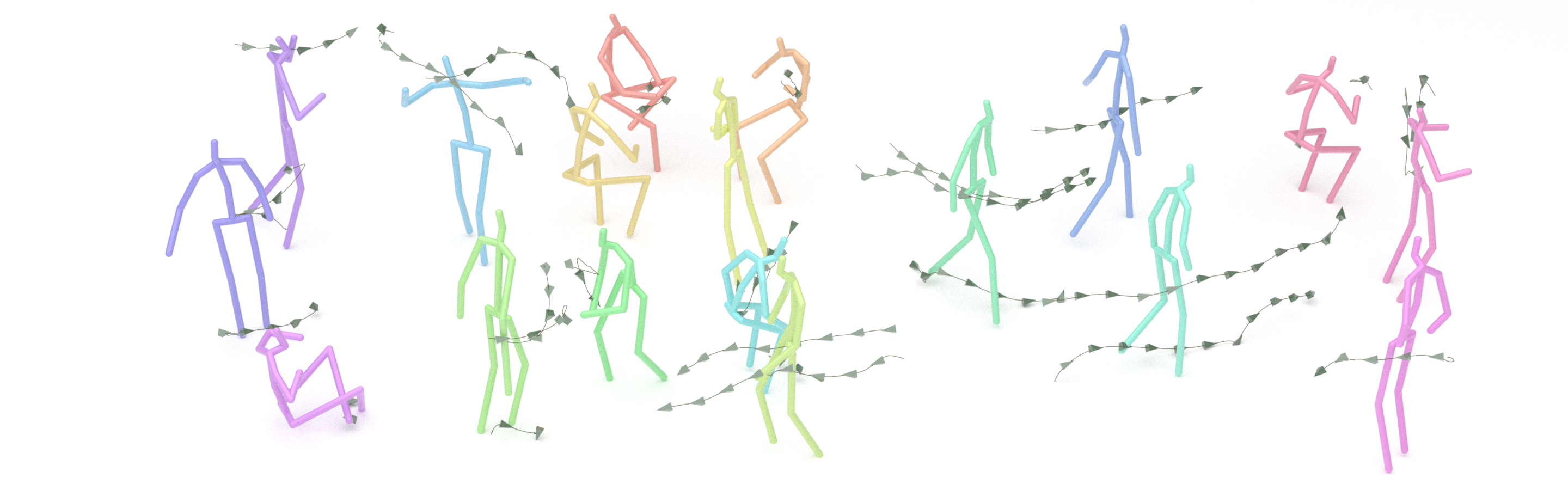}
    \captionof{figure}{Our method transforms 3D trajectories into a plausible human animation. The model requires only an initial pose and any number of input trajectories for each character in a scene. Generated animations can be composed to create vivid animated scenes.}
    \label{fig:teaser}
    \vspace{0.55em}
}

\makeatletter
\newcommand\notsotiny{\@setfontsize\notsotiny\@vipt\@viipt}
\makeatother

\begin{document}

\title{\textsc{TrajeVAE}: Controllable Human Motion Generation from Trajectories}

\author{Kacper Kania\textsuperscript{1}\\
\and
Marek Kowalski\textsuperscript{2}\\
\and
Tomasz Trzciński\textsuperscript{1,3,4}\\
\and
\textsuperscript{1}Warsaw University of Technology\\
\textsuperscript{2}Microsoft\\
\textsuperscript{3}Tooploox\\
\textsuperscript{4}Jagiellonian University\\
}

\maketitle
\iftoggle{cvprfinal}{
    \newcommand{\codelink}{\url{https://github.com/kacperkan/trajevae/}}
}{
    \newcommand{\codelink}{$\langle$hidden during the review process$\rangle$}
}

\newcommand{\trajevae}{\textsc{\mbox{TrajeVAE}}}

\begin{abstract}
    The creation of plausible and controllable 3D human motion animations is a long-standing problem that requires a manual intervention of skilled artists. Current machine learning approaches can semi-automate the process, however, they are limited in a significant way: they can handle only a single trajectory of the expected motion that precludes fine-grained control over the output. To mitigate that issue, we reformulate the problem of future pose prediction into pose completion in space and time where multiple trajectories are represented as poses with missing joints. We show that such a framework can generalize to other neural networks designed for future pose prediction. Once trained in this framework, a model is capable of predicting sequences from any number of trajectories. We propose a novel transformer-like architecture, \trajevae{}, that builds on this idea and provides a versatile framework for 3D human animation. We demonstrate that \trajevae{} offers better accuracy than the trajectory-based reference approaches and methods that base their predictions on past poses. We also show that it can predict reasonable future poses even if provided only with an initial pose.
\end{abstract}

\section{Introduction}
Creating realistic human animation is one of the key components in
robotics, game, and movie industries. Typically when working on an animation, 
the animator 
starts with defining a character's skeleton. Parts of this skeleton are manually created and defined to be in a specific relation such that each joint can influence the position of other joints. Together, these parts form a \textit{kinematic chain}. While these relations are helpful to maintain skeleton movement constraints, they are not sufficient to create realistic animation.  In fact, the generation of such animations requires manual key-framing of the joint positions throughout the sequence. That becomes quickly an unfeasible task for complex motions. 

The procedure can be aided with recent advances in machine learning. The automation of character animation is a long-standing problem with multiple solutions proposed, including those based on neural networks and probabilistic models~\cite{rose1998verbs,rose2001artist, taylor2007modeling,fragkiadaki2015recurrent,holden2015learning,holden2016deep}. The main goal of these methods is to generate sequences of joint positions given some conditioning information, \ie control signal. 
This control signal can be any partial future information, for example, the direction of the movement, speed, type of an action being performed, coordinates of a particular joint, or any combination of the above~\cite{henter2020moglow, pavllo2019modeling,chen2020dynamic, ghorbani2020probabilistic, holden2017phase, zhang2018mode, wang2019combining}. However, these methods can handle only simple motions of a body, such as walking, running, or side-stepping, and cannot infer fine-grained motions of each of the joints. In many cases, this formulation becomes impractical, if we wanted to simulate a crowd where characters perform vivid actions such as jumping, bending, or hand waving.%

To solve this issue, we propose a data-driven approach to train a model to handle a variable number of pieces of information, \textit{trajectories}, for human motion generation. A single \textit{trajectory} refers to a particular skeleton joint, \eg elbow, and specifies where that joint should be located in each time step. We can choose whether we want to specify trajectories of a few joints and leave the rest to be generated by the model or generate a more specific motion by adding more trajectories. This formulation introduces an unprecedentedly flexible framework that encompasses all previous approaches while offering adjustable control over the generated motion.

We formulate the problem of predicting future poses from trajectories as a \textit{pose completion} problem to achieve our goal. We trace the inspiration for that formulation to the evolutionary cognitive skills of humans who are able to \textit{hallucinate} \cite{johansson1973visual} the rest of a human body out of a few markers that exhibit a human motion. While a similar formulation was firstly used in \cite{hernandez2019human}, it was applied to a significantly different task of predicting future motions from past frames. Similarly, \cite{kaufmann2020convolutional} predicts missing poses in a sequence where only some of pose frames are given. However, we notice that without introducing a structured bias into the training of these methods, they fail at generating realistic poses even if we provide several trajectories. Moreover, they are deterministic by design at cannot generate multiple, diverse motions. 

Since we cast our problem as structured \textit{pose completion}, we leverage recent advancements in stochastic tensor completion \cite{zheng2019pluralistic} and show the application of that paradigm on a novel motion generation model which we call \trajevae{}. Thanks to our formulation, we achieve a desirable property that the accuracy of generated motions increases when more information is provided at the input. 
At the same time, our model can predict future poses even if no trajectory is given. 
In industrial applications, our method can generate full-body animations for automatically-tracked joints while naturally handling missing information if some of the joints are not seen by the model. \trajevae{} outperforms trajectory-based baselines and methods based on several past full body frames in terms of accuracy. We additionally show that our formulation can be adapted to existing methods targeting the defined task, thus improving their results.

We summarize our contributions as follows:
\begin{itemize}
    \item a simple and general training paradigm that enables controllable generation of future poses from a variable number of input information pieces,
    \item \trajevae{} --- the first generative model that predicts diverse poses from any number of input trajectories,
    \item empirical study showing that our formulation can be successfully applied to existing methods for generating motion from a single trajectory to improve their results and enable them to use multiple trajectories.
\end{itemize}

\begin{figure*}
    \centering
    \includegraphics[width=\linewidth]{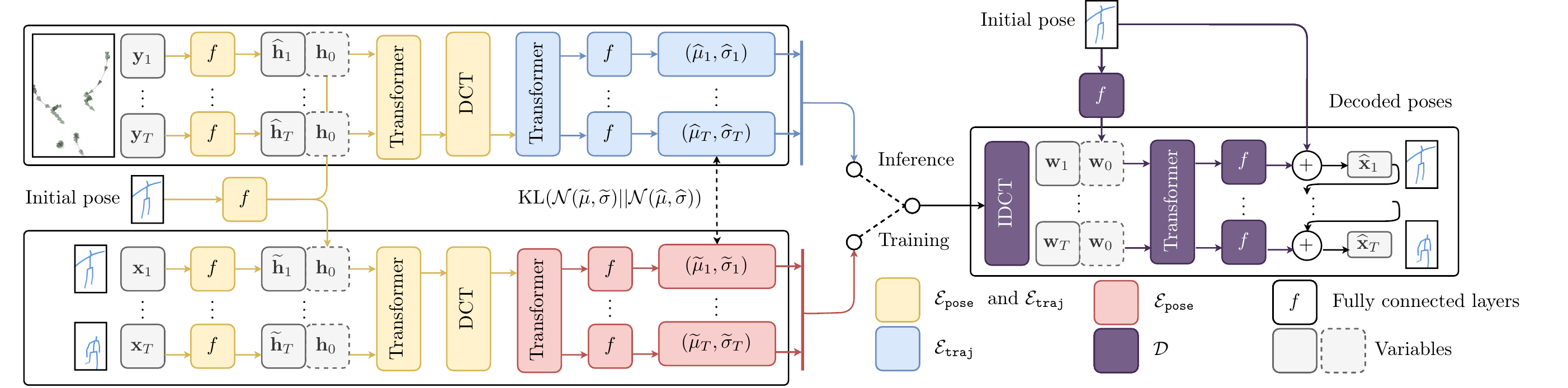}
    \caption{\textbf{Pipeline} of our \trajevae{} architecture trained in the \textit{pose completion} paradigm. \trajevae{} completes the sequence of missing joints and allows us to generate diverse poses from a variable number of input trajectories. During inference, we drop the part responsible for encoding poses $\{\mathbf{x}\}^T_{t=1}$.}
    \label{fig:pipeline}
    \vspace{-0.5em}
\end{figure*}

\section{Related Works}
\paragraph{Deterministic motion prediction} In recent years, multiple methods were proposed for predicting a single future motion based on a corresponding past sequence of poses~\cite{aksan2019structured, aksan2020attention, cai2020learning, cui2020learning, ghosh2017learning, gui2018adversarial, gopalakrishnan2019neural, gurumurthy2017deligan,li2018convolutional,li2020dynamic,mao2019learning,mao2020history,martinez2017human,pavllo2018quaternet,pavllo2019modeling,tang2018long} or video frames with missing poses \cite{martinez2017human, fragkiadaki2015recurrent,pavllo20193d, zhang2019predicting, kanazawa2019learning, iskakov2019learnable, zhang2021learning}. Cai~\etal~\cite{cai2020learning} and Aksan~\etal~\cite{aksan2020attention} use a transformer-like architecture to achieve this goal. Mao~\etal~\cite{mao2019learning,mao2020history} extend the pose representation by performing Discrete Cosine Transform (DCT) \cite{ahmed1974discrete} on joint coordinates. They additionally applied Graph Convolutional Networks (GCN) to incorporate the spatial information relationships between joints. Lebailly~\etal~\cite{lebailly2020motion} adapted inception modules \cite{szegedy2015going} to handle different temporal resolutions of the data. Kaufmann~\etal~\cite{kaufmann2020convolutional} uses a U-Net architecture to complete missing poses in the input tensor. Similar to our approach, Ruiz~\etal~\cite{hernandez2019human} treats the motion prediction as a pose completion problem. However, these  consider only randomly incomplete data while we train our method explicitly to leverage a variable number of available joint trajectories. While being successful, these methods are limited to predicting a single future pose sequence.

\paragraph{Stochastic motion prediction} To model the distribution of possible motions, recent works \cite{yan2018mt, walker2017pose, harvey2018recurrent, lin2018human, hernandez2019human, aliakbarian2019contextually, yuan2020dlow, kundu2019bihmp, kaufmann2020convolutional, jang2020constructing} leverage advances for generative modeling and build upon models such as Generative Adversarial Networks (GANs)~\cite{goodfellow2014generative}, Variational Autoencoders (VAEs)~\cite{kingma2013auto} or normalizing flows \cite{rezende2015variational}. These models enable sampling multiple future pose sequences and to accomplish this, they are often built as conditional models (CGANs~\cite{goodfellow2014generative} and CVAEs~\cite{kingma2013auto}). Barsoum~\etal~\cite{barsoum2018hp} use Wasserstein GAN~\cite{arjovsky2017wasserstein} in a sequence-to-sequence framework. To make poses more realistic, they regularize bone lengths and deviations between poses in consecutive frames. Walker~\etal~\cite{walker2017pose} apply VAE for the same goal and uses predicted poses to generate structurally consistent images.  Zhang~\etal~\cite{zhang2021learning} consider the motion generation given an action label to performed by the generated character. The authors also consider a transformer architecture to be suitable for time sequence modeling. While having high generation accuracy, these methods do not provide fine-grained control over generated outputs.

\paragraph{Stochastic prediction from trajectories} Existing methods \cite{pavllo2018quaternet,pavllo20193d,pavllo2019modeling,henter2020moglow,holden2017phase, chen2020dynamic,starke2019neural,zhang2018mode,dvorovzvnak2020monster} methods handle a single trajectory often represented as a target pelvis coordinate projected onto the floor's plane, or as a tuple of velocities for each axis. Pavllo~\etal~\cite{pavllo2018quaternet} encode a control signal in the form of the desired trajectory. Henter~\etal~\cite{henter2020moglow} input to the model a $t$ control signal and past control signals to predict $t+1$ pose. Holden~\etal~\cite{holden2017phase} split the motion into phases and model them with a phase-conditioned neural network. Yet these approaches are designed for a single trajectory and cannot generate motions like jumping jacks or hand waving where multiple trajectories need to be specified.

\paragraph{Image completion} As our method draws inspiration for image completion literature, we briefly summarize recent advanced in that field. 

Image inpainting~\cite{bertalmio2000image} is a well-known problem in computer vision. We posit that several of the already proposed approaches \cite{liu2018image,cai2020piigan,nazeri2019edgeconnect,xiong2019foreground,yu2018generative} can be successfully applied for pose completion where only a part of the pose is given. This way, we can leverage principles of these methods to improve general results. Liu~\etal~\cite{liu2018image} defines a partial convolution where the image is convolved only over pixels that are available in the input. Yu~\etal~\cite{yu2018generative} incorporates generative adversarial networks~\cite{goodfellow2014generative} and an attention mechanism to improve overall results. Zheng~\etal~\cite{zheng2019pluralistic} define a probabilistic model in the VAE framework that allowed the authors to generate diverse and realistic image completions.

\section{Method}
We introduce a novel paradigm of trajectory representation that enables fine-grained motion control with an arbitrary number of joint trajectories. We show that generating full-body poses from trajectories can be treated as a \textit{pose completion} problem. Then, we introduce \trajevae{} that builds on the paradigm and allows us to sample multiple, diverse poses which follow the conditioning trajectories. Throughout the whole paper, we parametrize trajectories and poses in 3D global coordinates. 
We show the overview of our model in \cref{fig:pipeline}.

\subsection{Handling multiple trajectories} We formulate the problem of predicting poses that follow a particular trajectory as a pose completion problem.  We denote a trajectory $\mathbf{Y}=\{\mathbf{y}_1, \dots, \mathbf{y}_T\}$ of length $T$ as vectors with $k \leq J$ known joint positions from a corresponding pose sequence $\mathbf{X}=\{\mathbf{x}_1, \dots, \mathbf{x}_T\}$, where $\mathbf{x}_t,\mathbf{y}_t\in\mathbb{R}^{3J}$. The trajectories of the unknown $J-k$ joints in $\mathbf{Y}$ are set to $0$.  The goal of the \textit{pose completion} task is to predict $\mathbf{X}$ given $\mathbf{Y}$. 

To mimic real-life scenarios, at training-time we randomly mask-out some of the joint the input pose sequence.
In this way, we reproduce the typical use cases such as occlusions or omissions of the animation artist. At each time step, we sample a matrix $\mathbf{M} \in \{0,1\}^{T\times 3J}$ that masks the same joints across $T$ time steps. Therefore, trajectories are obtained as $\mathbf{Y} = \mathbf{X} \odot \mathbf{M}$ where $\odot$ is the element-wise multiplication. 

We motivate the introduced paradigm as follows. Firstly, masking the poses in a principled, \textit{structured} way introduces a \textit{structural} bias into the model. The bias aids the model in learning particular distribution of poses. Such a model can outperform previous approaches such as \cite{kaufmann2020convolutional} by a significant margin even for a single trajectory. 
Secondly, thanks to that bias and in contrast to all related works that are limited to a single trajectory, our method allows the user to select how many trajectories are supplied to the network. 
As we show in the experiments, this paradigm enables a neural network model to handle a varying number of input trajectories. 

\subsection{Pose completion with a neural network}
To show the applicability of the introduced framework, we design \trajevae{} --- a Conditional Variational Autoencoder (CVAE) \cite{kingma2013auto} with a transformer-like architecture \cite{vaswani2017attention,wang2019t}, and a learnable prior distribution. %
The transformer architecture allows us to generate a sequence of poses in  parallel while the learnable prior increases the sampling diversity and plausibility.
Our model is an autoencoder with two encoders $\mathcal{E}_\code{pose}, \mathcal{E}_\code{traj}$ and a single decoder $\mathcal{D}$, where most of the blocks are shared. 
$\mathcal{E}_\code{pose}$ produces parameters of the posterior distribution $q_\code{pose}$ from the input poses $\mathbf{X}$. These parameters are optimized to match the trajectory prior distribution $p_\code{traj}$ parametrized by $\mathcal{E}_\code{traj}$.  Since we train the model to match distributions $q_\code{pose}$ and $p_\code{traj}$, the decoder $\mathcal{D}$ produces results that are similar for both $\mathcal{E}_\code{traj}$ and $\mathcal{E}_\code{pose}$.
During inference, we do not have access to the ground truth of a complete pose sequence and hence we drop $\mathcal{E}_\code{pose}$ that is responsible for encoding poses during training. 

\paragraph{Encoding poses and trajectories}
To take advantage of the similarity between representations of poses and trajectories, parameters of $\mathcal{E}_\code{traj}$ and $\mathcal{E}_\code{pose}$ are shared unless stated otherwise. We firstly encode the 3D coordinates of $\{\mathbf{y}_t\}_{t=1}^{T}$ and $\{\mathbf{x}_t\}_{t=1}^{T}$ with $\mathcal{E}_\code{traj}$ or $\mathcal{E}_\code{pose}$ to get $\{\hat{\mathbf{h}}_t\}^T_{t=1}$ and $\{\bar{\mathbf{h}}_t\}^T_{t=1}$ respectively.
We concatenate them with the initial pose representation $\mathbf{h}_0$ obtained from a separate neural network. $\hat{\mathbf{H}} = \{[\hat{\mathbf{h}}_t; \mathbf{h}_0]\}^T_{t=1}$ for trajectories and $\bar{\mathbf{H}} = \{[\bar{\mathbf{h}}_t; \mathbf{h}_0]\}^T_{t=1}$ for poses are passed to two self-attention layers~\cite{vaswani2017attention}. 
Then, we apply the Discrete Cosine Transform (DCT) for each feature in all vectors independently, and obtain vectors $\text{DCT}(\hat{\mathbf{H}})\in \mathbb{R}^{T\times (|\hat{\mathbf{h}}| + |\mathbf{h}_0|)}, \text{DCT}(\bar{\mathbf{{H}}}) \in \mathbb{R}^{T\times (|\bar{\mathbf{h}}| + |\mathbf{h}_0|)}$ in the frequency domain. As shown in \cite{zhang2020we}, most the variability of in the pose distribution concentrates in early components of DCT. Hence, sampling from the normal distribution in frequency domain increases diversity of generated poses.

Up to this point, all parameters for processing trajectories $\mathbf{Y}$ and poses $\mathbf{X}$ are shared to use the fact that trajectories represent masked future poses. We then split the pipeline into two unshared parts --- one for trajectories and one for poses --- that are composed of transformer-like encoders that facilitate information sharing between the latent codes.  The final multilayer perceptrons produce parameters $\left(\{\hat{\bm{\mu}}\}_{t=1}^T, \{\hat{\bm{\sigma}}\}_{t=1}^T\right)$ of a normal distribution for trajectories, and $\left(\{\bar{\bm{\mu}}\}_{t=1}^T, \{\bar{\bm{\sigma}}\}_{t=1}^T\right)$ for poses.

\paragraph{Learnable prior}
Constraining the latent space to a standard normal distribution $\mathcal{N}(0, I)$ as in VAEs is too restrictive and impedes the diversity of generated samples significantly. To overcome the problem and to provide a more flexible distribution, we make the prior learnable \cite{denton2018stochastic, zheng2019pluralistic} and define it as $p_\code{traj}(\hat{\mathbf{z}}_t | \mathbf{y}_1, \dots, \mathbf{y}_T)$. During training, we match the posterior distribution $q_\code{pose}(\bar{\mathbf{z}}_t | \mathbf{x}_1, \dots, \mathbf{x}_T)$ by optimizing the Kullback-Leibler divergence: 
\begin{equation*}
    -\text{KL}(q_\code{pose}(\bar{\mathbf{z}}_t | \mathbf{x}_1, \dots, \mathbf{x}_T)||p_\code{traj}(\hat{\mathbf{z}}_t | \mathbf{y}_1, \dots, \mathbf{y}_T)),
\end{equation*}
where $\hat{\mathbf{z}}_t \sim \mathcal{N}(\hat{\bm{\mu}}_t, \hat{\bm{\sigma}}_t)$ and $\bar{\mathbf{z}}_t \sim \mathcal{N}(\bar{\bm{\mu}}_t, \bar{\bm{\sigma}}_t)$  are samples from the prior and posterior distributions respectively.

\paragraph{Decoding poses}
We transform latent vectors of the poses $\{\bar{\mathbf{z}}\}^T_{t=1}$ during training and trajectory latent vectors $\{\hat{\mathbf{z}}\}^T_{t=1}$ during inference into the original time domain $\{\mathbf{w}\}_{t=1}^T$ with Inverse Discrete Cosine Transform (IDCT) \cite{ahmed1974discrete}. 
We additionally encode the initial pose with an MLP to obtain $\mathbf{w}_0$ as we found it improves overall results. A set of concatenated vectors
$\{[\mathbf{w}_t; \mathbf{w}_0]\}^T_{t=1}$ is decoded with a self-attention
decoder. The final fully connected layers predict offsets $\hat{\mathbf{o}}_t$
of the reconstructed pose $\hat{\mathbf{x}}_{t-1}$ from the time step $t-1$.
Finally, the reconstructed pose $\hat{\mathbf{x}}_t$ in the time step $t$ is obtained as:
\begin{equation}
    \hat{\mathbf{x}}_t = \sum_{\tau=1}^{t}\hat{\mathbf{x}}_{\tau -1} + \hat{\mathbf{o}}_\tau, \qquad \hat{\mathbf{x}}_0 = \mathbf{x}_0
\end{equation}
where $\mathbf{x}_0$ is the initial pose. 

Our approach decodes offsets of joints in step $t$ with respect to the pose $t-1$ without the need to access that pose. Therefore, the offsets can be predicted in parallel, and the final poses are obtained by a simple aggregation of offsets and adding them to the initial pose $\mathbf{x}_0$. This approach contrasts with the current notion of applying fully autoregressive decoders \cite{zhang2020we, yuan2020dlow, henter2020moglow} which suffer from slow inference. 

\paragraph{Training}  We train our \trajevae{} to accurately reconstruct poses, while maintaining the
posterior distribution close to the prior. We achieve this by 
optimizing the following objective \cite{kingma2013auto}:
\begin{equation}
    \label{eq:objective}
    \mathcal{L} = \mathcal{L}_\code{MSE} + \mathcal{L}_\code{KL},
\end{equation}
where $\mathcal{L}_\code{MSE}$ is the reconstruction term expressed as mean squared error:
\begin{equation}
    \label{eq:mse}
    \mathcal{L}_\code{MSE} = \sum_{t=1}^T\left|\left|\hat{\mathbf{x}}_t - \mathbf{x}_t\right|\right|^2_2,
\end{equation}
and $\mathcal{L}_\code{KL}$ keeps the posterior distribution close to the learnable prior by minimizing Kullback-Leibler divergence:
\begin{equation}
    \label{eq:kl}
    \mathcal{L}_\code{KL} = -\sum_{t=1}^T\beta\text{KL}(q_\code{pose}(\bar{\mathbf{z}}_t | \mathbf{X}, \mathbf{Y})||p_\code{traj}(\hat{\mathbf{z}}_t | \mathbf{Y})) .
\end{equation}

\paragraph{Masking future poses and data augmentation} Providing target poses $\{\mathbf{x}\}_{t=1}^T$
during training directly leads to overfitting and makes the network
unable to match the posterior with the prior. This further degrades the quality of reconstructed poses during inference. To overcome the problem, we mask input poses $\mathbf{X}$ with the inverse mask $\mathbf{M}$, that was used to obtain trajectories, as $\mathbf{X} \odot \left(1 - \mathbf{M}\right)$. 
This way, the pose encoder $\mathcal{E}_\code{pose}$ is
forced to leverage the information from the prior distribution of trajectories
which are structurally complementary to masked future poses.

\section{Experiments}

We evaluate \trajevae{} in two scenarios. Firstly, we evaluate the
performance of our method and of several baselines when we progressively add conditioning trajectories in the input. Secondly, we compare our method with
recent methods for a stochastic human generation. 
We also perform an ablation study to validate our design decisions.

\paragraph{Datasets} 
All our experiments are based on the Human3.6m dataset \cite{ionescu2013human3}.
It consists of 3.6 million video frames of 11 subjects performing 15 actions. 
We follow the evaluation protocol used in \cite{yuan2020dlow} and hence we use 17-joint poses. The training was done on subjects S1, S5, S6, S7, S8 while subjects S9 and S11 are left for the testing. We test the baselines and our method by predicting 2 seconds of future motion.

We represent human poses and trajectories as a set of joints parametrized as global coordinates. We also normalize each sequence such that the pelvis of the initial pose is located at the $(0, 0, 0)$ coordinate.
\begin{figure*}[t!]
    \centering
    \includegraphics[width=0.95\linewidth]{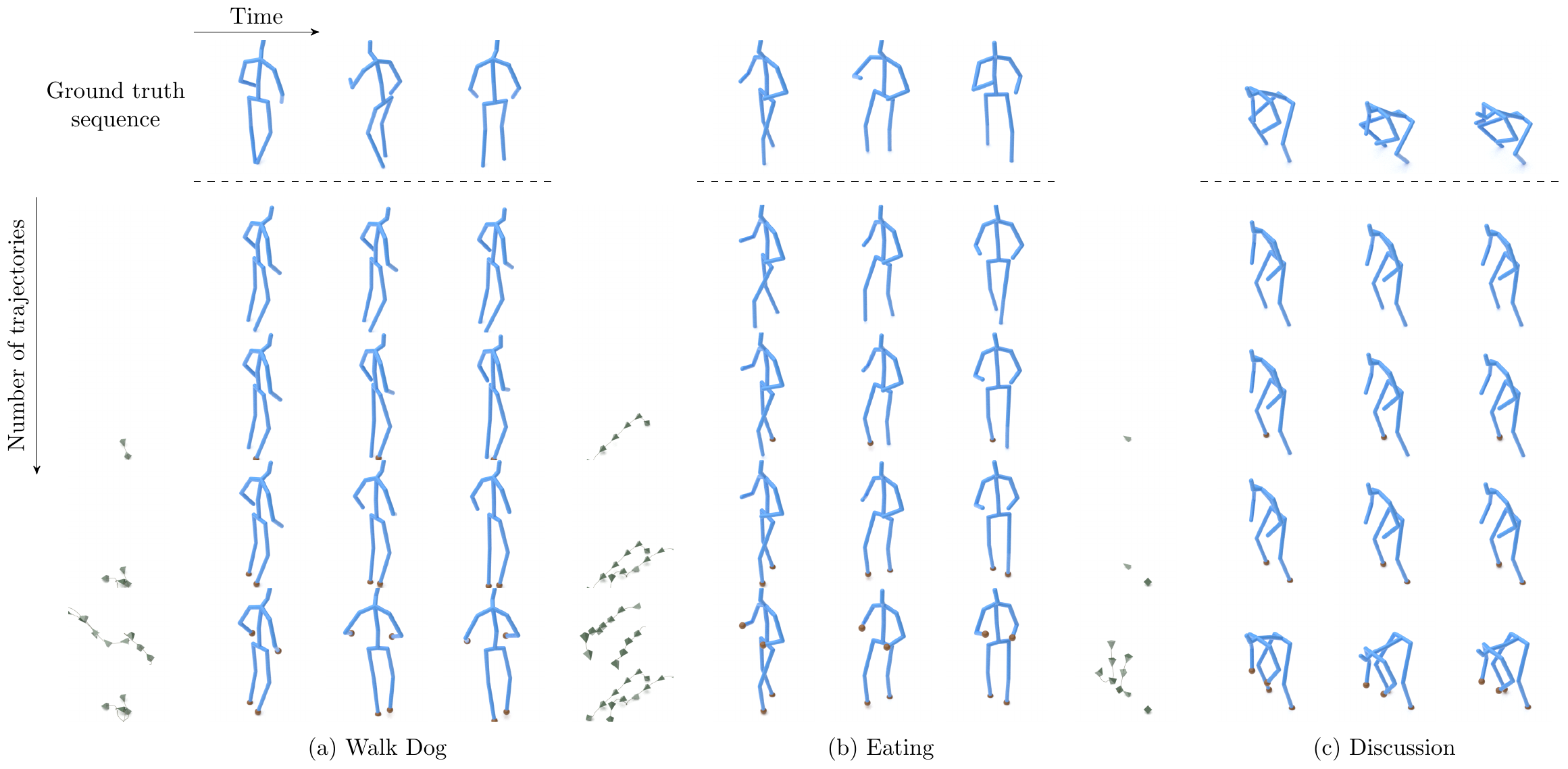}
    \caption{By using \trajevae{}, we can provide more trajectories in the
    input to create a realistic pose that follows a particular path. Joints that are not described by a trajectory are completed by our method. Here we show three sequences of motions generated by our method.  The joints that have a corresponding input trajectory are depicted as brown spheres. The top row shows the ground truth sequence. Rows below show generated sequences when more trajectories are given, in the order: right foot, left foot, right and left hands. Labels refer to classes in the Human3.6m dataset \cite{ionescu2013human3}.}
    \label{fig:sequences}
    \vspace{-0.5em}
\end{figure*}

\paragraph{Baselines}
To show the generality of our approach, we define baselines that, thanks to our paradigm, enable
generation of high quality human motions for a variable number of trajectories. We compare \trajevae{} with two approaches that are the most similar to ours and also use the principle of recovering the missing joints: MotionGAN \cite{hernandez2019human} and Motion Infilling \cite{kaufmann2020convolutional}. We show that using both of them out-of-the-box is not sufficient to produce high quality pose sequences from trajectories. Note also that these models are designed to be deterministic  and cannot produce diverse motions.

We also compare to a basic CVAE-RNN based on \cite{yuan2020dlow} and an adapted version of MoGlow \cite{henter2020moglow}. MoGlow in its original version supports only walking, running, and stepping motions. The conditioning signal used by the authors is expressed in terms of axis velocities for the pelvis joint. In its basic form, MoGlow can handle only a single trajectory to predict the motion. We provide additional implementation details of these baselines in the supplementary material.

In the second experiment, we follow the evaluation protocol of DLow \cite{yuan2020dlow} and use its baselines for the Human3.6M \cite{ionescu2013human3} dataset. In contrast to DLow however, our method requires only a single past frame (the initial pose) while the evaluation used by the authors assumed 25 past frames for Human3.6m and it does not allow to control predicted future motions.

\paragraph{Metrics}
We evaluate the methods using the \textit{diversity} and \textit{accuracy} metrics defined in \cite{yuan2020dlow}. 
    \metric{Average Pairwise Distance (APD)} describes the diversity of a set of size $K$ of motions sampled given the same input trajectory. It is expressed as the average $L_2$ distance between all pairs of generated motions \mbox{$\frac{1}{K(K-1)}\sum^K_{i=1}\sum^K_{j\neq i}||\hat{\mathbf{x}}_i - \hat{\mathbf{x}}_j||_2$}. 
    \metric{Average Displacement Error (ADE)} measures the accuracy of the reconstructed motion and calculates the average $L_2$ distance across all time steps between the ground truth motion and the motion from a generated set of $K$ motions that is the closest to the ground truth \mbox{$\frac{1}{T}\min_{\hat{\mathbf{x}} \in \hat{\mathbf{X}}}\sum_{t=1}^T||\hat{\mathbf{x}}_t - \mathbf{x}_t||_2$}. 
    \metric{Final Displacement Error (FDE)} calculates the $L_2$ distance between the pose in the last time step of ground truth motion and the motion from a generated set of $K$ motions that is the closest to the ground truth \mbox{$\min_{\hat{x} \in \hat{\mathbf{X}}}||\hat{\mathbf{x}}_T - \mathbf{x}_T||_2$}. 
    \metric{Multi-Modal ADE (MMADE)} and \metric{Multi-Modal FDE (MMFDE)} calculates an average of ADE and FDE respectively between a predicted motion and all samples in a cluster of motion sequences. We group these motions where the $L_2$ distance between their initial poses differs by less than~$\epsilon$.  

\paragraph{Implementation details}
At the training time, we obtain masks $\mathbf{M} \supset \mathbf{m}\in \{0, 1\}^{3J}$ by sampling from the Bernoulli
distribution $\mathcal{B}(p_\mathbf{m})$ with a probability $p_\mathbf{m}$. Then, we replicate the $\mathbf{m}$ vector $T$ times to create the structured mask $\mathbf{M} \in \{0, 1\}^{T \times 3J} $. 
We set $p_\mathbf{m} = 0.85$ so that the network sees 3 -- 4 trajectories on average in the input. We motivate that number to be a sensible trade-off between the accuracy and effort of defining trajectories when using \trajevae{} in practice.

We train \trajevae{} and the corresponding baselines with the Adam optimizer
\cite{kingma2014adam} with learning rate set to $0.0001$ and multiplied by $0.25$ every
$80,000$ training steps. We set $\beta=0.01$ in the KL term, the batch size $= 64$ and we train models for $240,000$ steps.

\subsection{Qualitative results}

\paragraph{Reconstructed sequences} To visually examine the proposed method, we generate a set of poses while
changing the number of input trajectories. \cref{fig:sequences} shows individual frames from selected animation sequences (refer to the supplementary material to see full video clips). When more trajectories are provided, the generated sequence resembles the ground truth more closely. However, even if no trajectory is provided, \trajevae{} generates
plausible poses. We achieve this by providing the initial pose to the model during the decoding phase. We note that the initial pose heavily biases the model due to the nature of the dataset.

\begin{figure}[t!]
    \centering
    \includegraphics[width=0.95\linewidth]{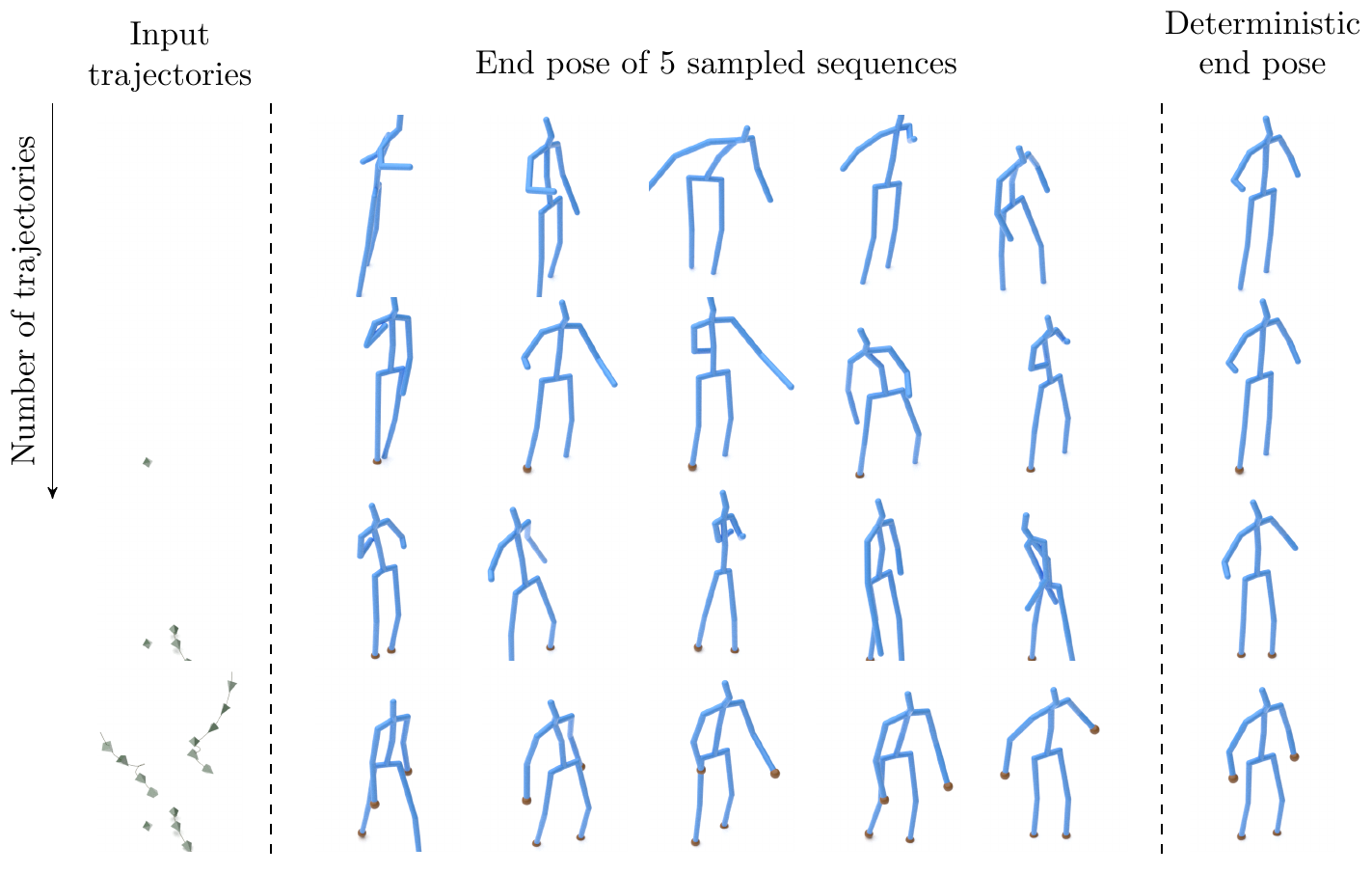}
    \caption{\trajevae{} allows us to use a variable number of trajectories and sample multiple diverse pose sequences. We show in each row input trajectories, end poses for 5 sampled sequences, and the end pose for a sequence decoded from trajectory means $\{\hat{\mu}\}^T_{t=1}$. The joints that have a specified trajectory are colored in brown.}
    \label{fig:sample-vs-joints}
    \vspace{-0.5em}
\end{figure}
\paragraph{Diversity vs.~number of input trajectories} Since \trajevae{} allows us to sample latent DCT components from a learnable prior distribution, we can generate multiple, diverse samples for the same set of conditioning trajectories. We show the last frames of such generated samples in \cref{fig:sample-vs-joints}. When no trajectories are present, the method generates the most diverse outputs, while retaining the plausibility of poses. As we noticed by examining the generated sequences, such poses can represent waving, bending, or dancing-like motions. When four trajectories are given, generated poses converge towards the ground truth. 

Due to the MSE term in \cref{eq:mse}, the model is not forced to exactly reproduce the trajectories, and therefore the results plateau when we provide more than ten trajectories. Applying $L_1$ loss instead of $L_2$ mitigates that issue but significantly impedes the diversity of generated samples.

\begin{figure}[t!]
    \centering
    \includegraphics[width=0.95\linewidth]{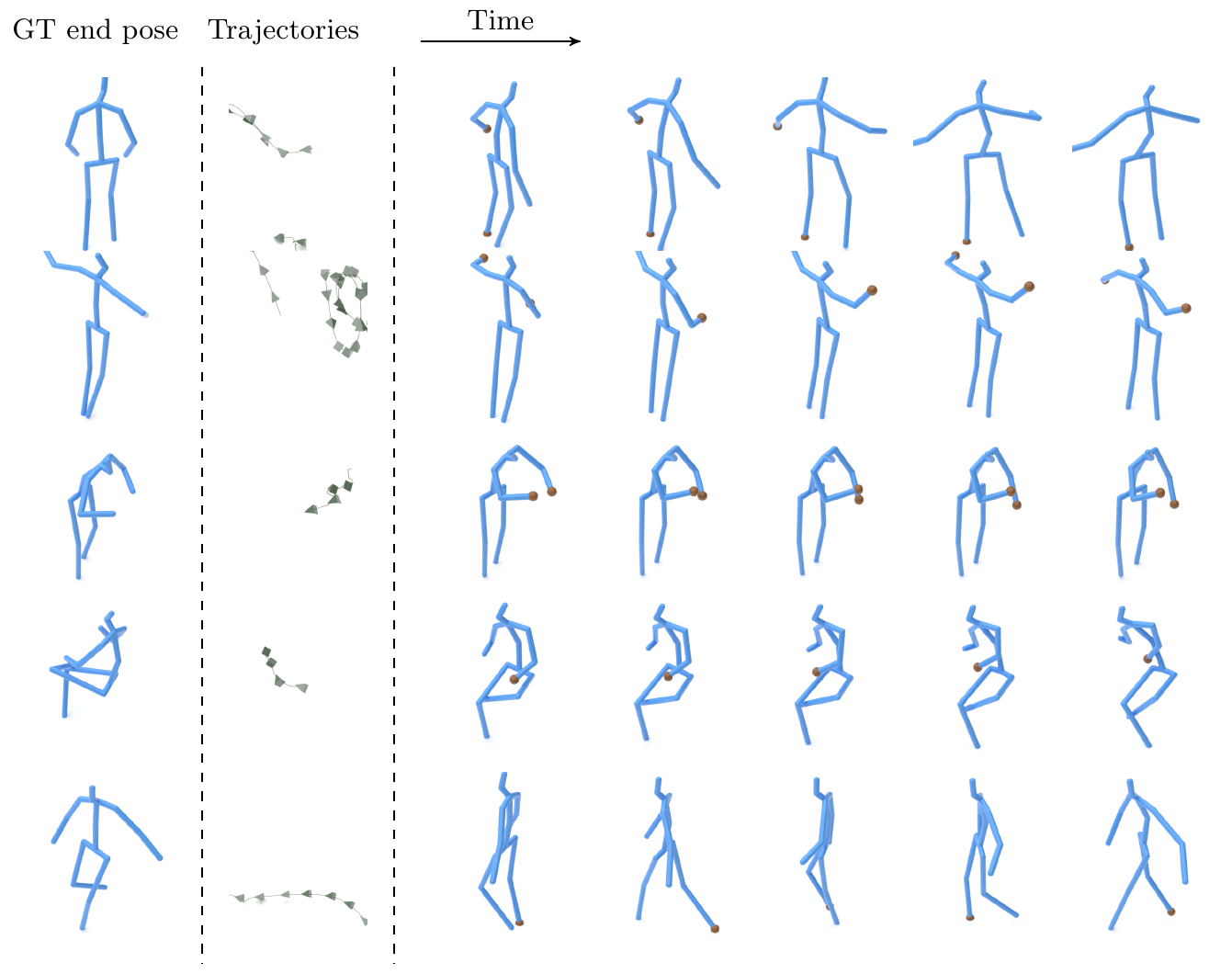}
    \caption{\textsc{TrajeVAE} is not limited to only walking, running, or standing, as are other related methods, and can be applied to any motion type. Each row represents a generated sequence for a different motion class given specific trajectories. The joints that have a specified trajectory are colored in brown.}
    \label{fig:generalization}
\end{figure}

\paragraph{Generalization} Finally, we empirically show that, in contrast to other works on controllable human motion generation, \trajevae{} can be applied to any set of motions, \eg dancing, sitting, waving, and others. As we present in \cref{fig:generalization}, \trajevae{} generalizes to a variety of different sets of motions that were not handled by previous methods. We additionally show in the supplementary, that given the same trajectory but a different initial pose, our method still generates poses that follow the provided trajectory. It confirms that our approach leverages both the initial pose and trajectories, and can be used to generate animations beyond the ones found in the dataset.

\subsection{Quantitative Results}
\begin{table}[t!]
\centering
\notsotiny
\begin{tabular}{lcccccc}
    \toprule
             \multirow{2}{*}{Method} & \multirow{2}{*}{$k$} & APD  & ADE  & FDE  & MMADE  & MMFDE  \\ 
         & & $\uparrow$ & $\downarrow$ & $\downarrow$& $\downarrow$& $\downarrow$\\
    \midrule
                             \textbf{\trajevae{}} & \multirow{5}{*}{0}   &          8.462 & \textbf{0.518} & \textbf{0.678} & \textbf{0.596} & \textbf{0.703} \\
              MotionGAN \cite{hernandez2019human} &   &            0.0 &          1.174 &          1.274 &          1.174 &          1.270 \\
Motion Infilling \cite{kaufmann2020convolutional} &   &            0.0 &          1.203 &          1.209 &          1.197 &          1.207 \\
                   MoGlow \cite{henter2020moglow} &   &          1.786 &          0.548 &          0.776 &          0.626 &          0.803 \\
                                         CVAE-RNN &   & \textbf{9.579} &          0.611 &          0.723 &          0.639 &          0.724 \\
    \midrule
                             \textbf{\trajevae{}} &  \multirow{5}{*}{1} &          6.641 & \textbf{0.463} & \textbf{0.602} & \textbf{0.581} & \textbf{0.672} \\
              MotionGAN \cite{hernandez2019human} &   &            0.0 &          0.936 &          1.120 &          0.958 &          1.147 \\
Motion Infilling \cite{kaufmann2020convolutional} &   &            0.0 &     $>10^4$ &     $>10^4$ &     $>10^4$ &     $>10^4$ \\
                   MoGlow \cite{henter2020moglow} &   &          1.813 &          0.546 &          0.773 &          0.625 &          0.801 \\
                                         CVAE-RNN &   & \textbf{9.498} &          0.603 &          0.733 &          0.647 &          0.753 \\
    \midrule
                             \textbf{\trajevae{}} &  \multirow{5}{*}{2} &          6.334 & \textbf{0.450} & \textbf{0.581} & \textbf{0.581} & \textbf{0.668} \\
              MotionGAN \cite{hernandez2019human} &   &            0.0 &          0.904 &          1.131 &          0.953 &          1.185 \\
Motion Infilling \cite{kaufmann2020convolutional} &   &            0.0 &     $>10^4$ &    $>10^4$ &     $>10^4$ &    $>10^4$ \\
                   MoGlow \cite{henter2020moglow} &   &          1.861 &          0.544 &          0.768 &          0.623 &          0.797 \\
                                         CVAE-RNN &   & \textbf{9.496} &          0.588 &          0.714 &          0.642 &          0.745 \\
    \midrule
                             \textbf{\trajevae{}} & \multirow{5}{*}{3}  &          5.037 & \textbf{0.375} & \textbf{0.488} & \textbf{0.579} & \textbf{0.664} \\
              MotionGAN \cite{hernandez2019human} &   &            0.0 &          0.812 &          1.063 &          0.948 &          1.220 \\
Motion Infilling \cite{kaufmann2020convolutional} &   &            0.0 &         22.727 &         76.233 &         22.801 &         76.326 \\
                   MoGlow \cite{henter2020moglow} &   &          1.844 &          0.540 &          0.766 &          0.623 &          0.798 \\
                                         CVAE-RNN &   & \textbf{9.309} &          0.516 &          0.626 &          0.614 &          0.699 \\
    \midrule
                             \textbf{\trajevae{}} & \multirow{5}{*}{4}  &          4.069 & \textbf{0.325} & \textbf{0.428} & \textbf{0.584} & \textbf{0.674} \\
              MotionGAN \cite{hernandez2019human} &   &            0.0 &          0.675 &          0.919 &          0.932 &          1.226 \\
Motion Infilling \cite{kaufmann2020convolutional} &   &            0.0 &          1.067 &          1.220 &          1.258 &          1.468 \\
                   MoGlow \cite{henter2020moglow} &   &          1.858 &          0.530 &          0.750 &          0.619 &          0.790 \\
                                         CVAE-RNN &   & \textbf{9.242} &          0.459 &          0.560 &          0.602 &          0.679 \\
    \bottomrule
\end{tabular}
\caption{Influence of the \textbf{number of trajectories} $k$ on the quantitative results. Note that both MotionGAN \cite{hernandez2019human} and Motion Infilling \cite{kaufmann2020convolutional} have deterministic architectures and cannot produce diverse motions.
Best results are in bold. }
\label{tab:controllability-experiment}                                
\vspace{-1.5em}
\end{table}                                          

\paragraph{Controlling future motion prediction}
We evaluate the quality of generated samples while changing the number of input trajectories. We first compute the metrics with no provided trajectories and then progressively increase their number by adding the following: right foot, left foot, right hand, and left hand. This order was motivated by the variance of coordinates of joints they correspond to, \ie, the joint with highest variance in the dataset was added first. 
In each case, we sample $K = 50$ poses to calculate metrics. Results are summarized in \cref{tab:controllability-experiment}. As expected, adding more trajectories decreases the diversity (APD) of samples since the pose is restricted to follow a particular path. At the same time, the accuracy (ADE) of generated samples improves. Our \trajevae{} obtains the best results in terms of the reconstruction quality in comparison to other methods. The higher diversity of CVAE-RNN is caused by its structure --- in contrast to \trajevae{}, CVAE-RNN encodes the whole sequence into a single latent code instead of multiple components. Therefore, two samples from the prior may have a significantly different structure in the output. However, such an architecture suffers from pose averaging where most of the pose frames are the same in a generated sequence \cite{fragkiadaki2015recurrent, li2017auto} which leads to the inferior accuracy of the reconstructed motions. We hypothesize that Motion Infilling \cite{kaufmann2020convolutional} obtains high error on the reconstruction due to two issues: it is trained with the $L_1$ reconstruction loss (which also worked detrimentally for \trajevae{}) and the simple masking scheme where each coordinate for all joints across the sequence is randomly masked out. 

\begin{figure}
    \centering
    \includegraphics[width=0.95\linewidth]{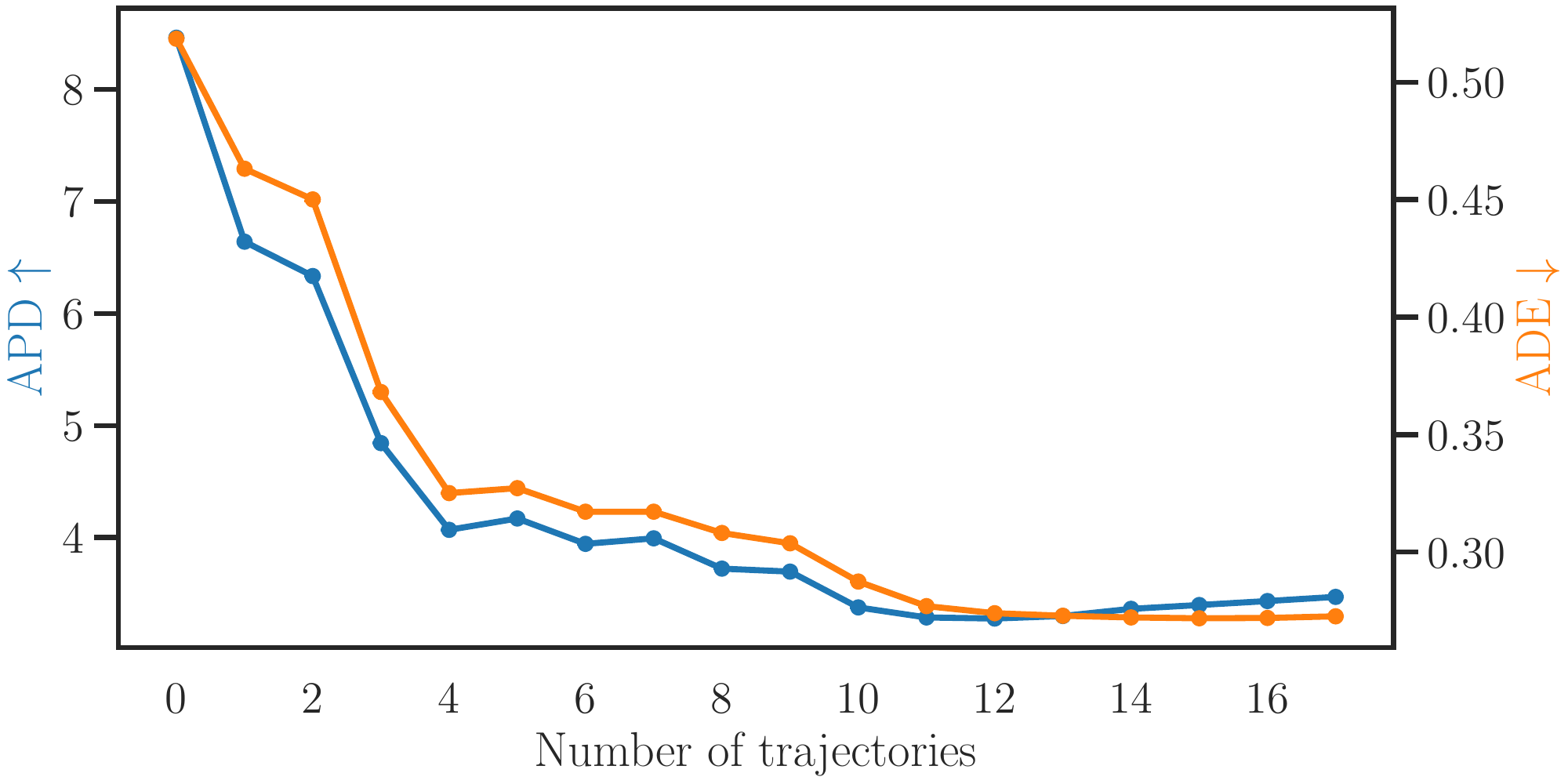}
    \caption{Providing more trajectories in the input causes that the reconstruction error (ADE) of the reconstruction decreases gradually for the cost of the decreased diversity (APD).}
    \label{fig:apd-vs-ade-trajectories}
\end{figure}

In \cref{fig:apd-vs-ade-trajectories}, we show that including more trajectories in the input improves the reconstruction quality. Notice that including too many trajectories plateaus the quality. We conclude that it is caused by the randomness in the latent space, and hence the network is unable to properly encode all the input trajectories to reconstruct poses accurately. Interestingly, the diversity (APD) is the lowest for ten trajectories and slightly increases when we add more trajectories. We argue that this phenomenon may be caused by the order we use when adding trajectories. Due to the low influence of the last seven trajectories on the sequence, they do not affect the quality.

\begin{table}[t!]
\centering
\scriptsize
\begin{tabular}{lccccc}
    \toprule
         \multirow{2}{*}{Method}  & APD  & ADE  & FDE  & MMADE  & MMFDE  \\ 
         & $\uparrow$ & $\downarrow$ & $\downarrow$& $\downarrow$& $\downarrow$\\
    \midrule             
        \textbf{\trajevae{}} ($ k = 0$) & 8.462 &  0.518 &  0.678 &  0.596 &  0.703 \\
        \textbf{\trajevae{}} ($k = 1$) &              6.641 &           0.463 &           0.602 &           0.581 &           0.672 \\
        \textbf{\trajevae{}} ($k = 2$) &              6.334 &           0.450 &           0.581 &           0.581 &           0.668 \\
        \textbf{\trajevae{}} ($k = 3$) &              5.037 &           0.375 &           0.488 &           0.579 &           0.664 \\
        \textbf{\trajevae{}} ($k = 4$) &              4.069 &  \textbf{0.325} &  \textbf{0.428} &           0.584 &           0.674 \\
    \midrule
        DLow \cite{yuan2020dlow} & \textbf{11.741} & 0.425 &  0.518 & \bf 0.495 & \bf 0.531 \\
        ERD \cite{fragkiadaki2015recurrent} & 0.0  & 0.722 & 0.969 & 0.776 & 0.995 \\
        acLSTM \cite{li2017auto} & 0.0  & 0.789 & 1.126 & 0.849 & 1.139 \\
        Pose-Knows \cite{walker2017pose}& 6.723  & 0.461 & 0.560 & 0.522 & 0.569  \\
        MT-VAE \cite{yan2018mt}& 0.403  & 0.457 & 0.595 & 0.716 & 0.883 \\
        HP-GAN \cite{barsoum2018hp}& 7.214  & 0.858 & 0.867 & 0.847 & 0.858 \\
        Best-of-Many \cite{bhattacharyya2018accurate} & 6.265  & 0.448 & 0.533 & 0.514 & 0.544 \\
        GMVAE \cite{dilokthanakul2016deep} & 6.769  & 0.461 & 0.555 & 0.524 & 0.566 \\
        DeLiGAN \cite{gurumurthy2017deligan} & 6.509  & 0.483 & 0.534 & 0.520 & 0.545 \\
        DSF \cite{yuan2019diverse} & 9.330  & 0.493 & 0.592 & 0.550 & 0.599 \\
    \bottomrule
\end{tabular}
\caption{\textbf{Quantitative results} for the Human3.6M dataset when $K=50$ samples are generated.
For our method, we assume different scenarios when \mbox{$k=\{0, 1, 2, 3, 4\}$} trajectories are provided. We explicitly delimit DLow~\cite{yuan2020dlow} and its baselines as these methods do not use trajectories and predict future poses from 25 past frames. 
Best results are in bold.}
\label{tab:base-experiment}                                
\end{table}               

\begin{table}[t!]
\centering
\scriptsize
\begin{tabular}{lcc}
    \toprule
         Method  &  ADE $\downarrow$ & FDE  $\downarrow$ \\
    \midrule             
        \textbf{\trajevae{}} ($ k = 0$) &  0.529 &  0.779 \\
        \textbf{\trajevae{}} ($k = 1$) & 0.491 & 0.735 \\
        \textbf{\trajevae{}} ($ k = 2$) &  0.477 &  0.710 \\
        \textbf{\trajevae{}} ($k = 3$) & 0.396 & 0.593\\
        \textbf{\trajevae{}} ($k = 4$) & \textbf{0.345} & \textbf{0.522} \\
    \midrule
        DLow \cite{yuan2020dlow} & 1.126 & 1.652 \\
    \bottomrule
\end{tabular}
\caption{\textbf{Quantitative results} for the Human3.6M dataset when a single future pose sequence ($K=1$) is generated. For our method, we assume different scenarios when \mbox{$k=\{0, 1, 2, 3, 4\}$} trajectories are provided. In this experiment, \trajevae{} decodes predicted means $\{\hat{\mu}\}^T_{t=1}$ of trajectories. Best results are in bold.}
\label{tab:single-sample-experiment}                                
\vspace{-0.75em}
\end{table}               

\paragraph{Comparison with other methods for future motion generation} We also compare \trajevae{} with methods that generate future poses from a set of past frames. 
We evaluate the methods in two scenarios: when sampling $K=50$ different poses, and when sampling only a single pose. We show results for \trajevae{} when $k=\{0, 1,2, 3, 4\}$ trajectories are present in the input. Results are summarized in \cref{tab:base-experiment} for $K=50$ and \cref{tab:single-sample-experiment} for $K=1$. When $K=50$, while working on an initial pose only without a conditioning trajectory, \trajevae{} produces poses comparable in terms of the accuracy to those generated by other approaches even though it was not trained specifically for this task. When we use $k = 4$ trajectories, the accuracy is better than for all other baselines. Note that both DLow~\cite{yuan2020dlow} and DSF~\cite{yuan2019diverse} were specifically trained to generate diverse samples while \trajevae{} implicitly provides high diversity by having a more flexible, learnable prior distribution. DLow is additionally constrained to always generate $K$ poses and cannot be extended to more samples.

When $K=1$, our method is capable of creating accurate pose sequences for a single sample. To adapt DLow to this scenario, we take the first pose from all its $K$ generated poses. DLow's much lower accuracy compared to experiments with $K=50$ is caused by its diverse outputs, many of which do no match the ground truth. Since DLow does not allow for controlling the trade-off between diversity and accuracy, it is unable to generate an accurate pose sequence reliably if only a single sample is generated. In comparison, our method is capable of controlling that trade-off and thus producing accurate results even if only a single output sample is generated.

\begin{table}[t!]
\centering
\scriptsize
\begin{tabular}{lccccc}
    \toprule
         \multirow{2}{*}{Method}  & APD  & ADE  & FDE  & MMADE  & MMFDE  \\ 
         & $\uparrow$ & $\downarrow$ & $\downarrow$& $\downarrow$& $\downarrow$\\
    \midrule             
    Base  &                      1.220 &           0.481 &           0.695 &           0.640 &           0.811 \\
    + Learnable prior &          1.749 &           \textbf{0.448} &           \textbf{0.622} &           0.618 &           0.753 \\
    + DCT  &                     \textbf{7.014} &           0.487 &           0.637 &           0.602 &           0.703 \\
    + Masked future poses &      6.803 &           0.472 &           0.623 &           \textbf{0.594} &           \textbf{0.695} \\
    \bottomrule
\end{tabular}
\caption{Influence of design decisions on obtained results for a single given trajectory (a right foot). Best results are in bold. For the ablation study with more trajectories, refer to the supplementary.}
\label{tab:ablation-study}                                
\vspace{-0.5em}
\end{table}                                          
\paragraph{Ablation study}
Finally, we perform an ablation study of our design decisions: making the prior distribution
learnable, using the Discrete Cosine Transform in the latent space, and masking future poses with the mask
$1 - \mathbf{M}$. We show obtained results in \cref{tab:ablation-study}.

The transformer-like architecture already obtains remarkable results. However, these can be improved by using learnable prior. While DCT reduces the accuracy, we found it to be a necessary component to obtain diverse samples.

We also found it beneficial to apply additional regularization technique by masking future poses with the mask $1 - \mathbf{M}$. By masking the poses, the network has to learn to encode more information from poses and stops to rely entirely on the prior distribution.

\section{Conclusions and limitations}
We introduced the notion of trajectory-conditioned pose generation as a pose completion problem. It allowed us to define \trajevae{} --- a method for controllable and stochastic human animation generation.
We showed that the paradigm of structured dropping of joints during training, creates a model that can generate realistic poses that follow an arbitrary number of trajectories. Obtained results show the applicability of our method in designing realistic human animations. While our approach trivially generalizes to other data representations, applying it to full-body parametric models, such as SMPL~\cite{pavlakos2019expressive, zhang2020we}, is of high importance. 

We identify two limitations of our approach. Firstly, generated poses do not follow the trajectories exactly. While we could resort to a cGAN~\cite{goodfellow2014generative} model as a possible solution for its unprecedented quality of generated samples~\cite{karras2019style}, the application of GANs to structured time series data is still a challenge.

Secondly, when the initial pose is ambiguous about what action it represents, \trajevae{} tends to stretch bones when we input only a single or none trajectories. Applying exponential maps~\cite{pavllo2018quaternet} constrains bone lengths but this would limit our method to only work with skeleton structures that have a clearly defined kinematic chain.

\section{Ethical concerns}
We do not identify immediate abuses of our approach in real world applications. \trajevae{} can be used to reanimate characters, however their skeletons still need to be manually defined. We regard the recent progress in neural radiance field \cite{mildenhall2020nerf} methods and their applications for human reposing \cite{su2021nerf} as a potential ethical concert.

\iftoggle{cvprfinal}{
\section*{Acknowledgements}
This work was supported by Microsoft Research through its EMEA PhD Scholarship Programme. We thank Eric Hedlin and members of CVLab at Warsaw University of Technology for insightful comment.
}{}

{
\small
\bibliographystyle{ieee_fullname}
\bibliography{egbib}

\begin{thebibliography}{10}\itemsep=-1pt

\bibitem{ahmed1974discrete}
Nasir Ahmed, T. Natarajan, and Kamisetty~R Rao.
\newblock Discrete cosine transform.
\newblock {\em IEEE transactions on Computers}, 100(1):90--93, 1974.

\bibitem{aksan2020attention}
Emre Aksan, Peng Cao, Manuel Kaufmann, and Otmar Hilliges.
\newblock Attention, please: A spatio-temporal transformer for 3d human motion
  prediction.
\newblock {\em arXiv preprint arXiv:2004.08692}, 2020.

\bibitem{aksan2019structured}
Emre Aksan, Manuel Kaufmann, and Otmar Hilliges.
\newblock Structured prediction helps 3d human motion modelling.
\newblock In {\em Proceedings of the IEEE/CVF International Conference on
  Computer Vision}, pages 7144--7153, 2019.

\bibitem{aliakbarian2019contextually}
Sadegh Aliakbarian, Fatemeh~Sadat Saleh, Lars Petersson, Stephen Gould, and
  Mathieu Salzmann.
\newblock Contextually plausible and diverse 3d human motion prediction.
\newblock {\em arXiv preprint arXiv:1912.08521}, 2019.

\bibitem{arjovsky2017wasserstein}
Martin Arjovsky, Soumith Chintala, and L{\'e}on Bottou.
\newblock Wasserstein generative adversarial networks.
\newblock In {\em International conference on machine learning}, pages
  214--223. PMLR, 2017.

\bibitem{barsoum2018hp}
Emad Barsoum, John Kender, and Zicheng Liu.
\newblock Hp-gan: Probabilistic 3d human motion prediction via gan.
\newblock In {\em Proceedings of the IEEE conference on computer vision and
  pattern recognition workshops}, pages 1418--1427, 2018.

\bibitem{bertalmio2000image}
Marcelo Bertalmio, Guillermo Sapiro, Vincent Caselles, and Coloma Ballester.
\newblock Image inpainting.
\newblock In {\em Proceedings of the 27th annual conference on Computer
  graphics and interactive techniques}, pages 417--424, 2000.

\bibitem{bhattacharyya2018accurate}
Apratim Bhattacharyya, Bernt Schiele, and Mario Fritz.
\newblock Accurate and diverse sampling of sequences based on a “best of
  many” sample objective.
\newblock In {\em Proceedings of the IEEE Conference on Computer Vision and
  Pattern Recognition}, pages 8485--8493, 2018.

\bibitem{cai2020piigan}
Weiwei Cai and Zhanguo Wei.
\newblock Piigan: Generative adversarial networks for pluralistic image
  inpainting.
\newblock {\em IEEE Access}, 8:48451--48463, 2020.

\bibitem{cai2020learning}
Yujun Cai, Lin Huang, Yiwei Wang, Tat-Jen Cham, Jianfei Cai, Junsong Yuan, Jun
  Liu, Xu Yang, Yiheng Zhu, Xiaohui Shen, et~al.
\newblock Learning progressive joint propagation for human motion prediction.
\newblock In {\em European Conference on Computer Vision}, pages 226--242.
  Springer, 2020.

\bibitem{chen2020dynamic}
Wenheng Chen, He Wang, Yi Yuan, Tianjia Shao, and Kun Zhou.
\newblock Dynamic future net: Diversified human motion generation.
\newblock In {\em Proceedings of the 28th ACM International Conference on
  Multimedia}, pages 2131--2139, 2020.

\bibitem{cui2020learning}
Qiongjie Cui, Huaijiang Sun, and Fei Yang.
\newblock Learning dynamic relationships for 3d human motion prediction.
\newblock In {\em Proceedings of the IEEE/CVF Conference on Computer Vision and
  Pattern Recognition}, pages 6519--6527, 2020.

\bibitem{denton2018stochastic}
Emily Denton and Rob Fergus.
\newblock Stochastic video generation with a learned prior.
\newblock In {\em International Conference on Machine Learning}, pages
  1174--1183. PMLR, 2018.

\bibitem{dilokthanakul2016deep}
Nat Dilokthanakul, Pedro~AM Mediano, Marta Garnelo, Matthew~CH Lee, Hugh
  Salimbeni, Kai Arulkumaran, and Murray Shanahan.
\newblock Deep unsupervised clustering with gaussian mixture variational
  autoencoders.
\newblock {\em arXiv preprint arXiv:1611.02648}, 2016.

\bibitem{dvorovzvnak2020monster}
Marek Dvoro{\v{z}}{\v{n}}{\'a}k, Daniel S{\`y}kora, Cassidy Curtis, Brian
  Curless, Olga Sorkine-Hornung, and David Salesin.
\newblock Monster mash: a single-view approach to casual 3d modeling and
  animation.
\newblock {\em ACM Transactions on Graphics (TOG)}, 39(6):1--12, 2020.

\bibitem{fragkiadaki2015recurrent}
Katerina Fragkiadaki, Sergey Levine, Panna Felsen, and Jitendra Malik.
\newblock Recurrent network models for human dynamics.
\newblock In {\em Proceedings of the IEEE International Conference on Computer
  Vision}, pages 4346--4354, 2015.

\bibitem{ghorbani2020probabilistic}
Saeed Ghorbani, Calden Wloka, Ali Etemad, Marcus~A Brubaker, and Nikolaus~F
  Troje.
\newblock Probabilistic character motion synthesis using a hierarchical deep
  latent variable model.
\newblock In {\em Computer Graphics Forum}, volume~39, pages 225--239. Wiley
  Online Library, 2020.

\bibitem{ghosh2017learning}
Partha Ghosh, Jie Song, Emre Aksan, and Otmar Hilliges.
\newblock Learning human motion models for long-term predictions.
\newblock In {\em 2017 International Conference on 3D Vision (3DV)}, pages
  458--466. IEEE, 2017.

\bibitem{goodfellow2014generative}
Ian~J Goodfellow, Jean Pouget-Abadie, Mehdi Mirza, Bing Xu, David Warde-Farley,
  Sherjil Ozair, Aaron Courville, and Yoshua Bengio.
\newblock Generative adversarial networks.
\newblock {\em arXiv preprint arXiv:1406.2661}, 2014.

\bibitem{gopalakrishnan2019neural}
Anand Gopalakrishnan, Ankur Mali, Dan Kifer, Lee Giles, and Alexander~G
  Ororbia.
\newblock A neural temporal model for human motion prediction.
\newblock In {\em Proceedings of the IEEE/CVF Conference on Computer Vision and
  Pattern Recognition}, pages 12116--12125, 2019.

\bibitem{gui2018adversarial}
Liang-Yan Gui, Yu-Xiong Wang, Xiaodan Liang, and Jos{\'e}~MF Moura.
\newblock Adversarial geometry-aware human motion prediction.
\newblock In {\em Proceedings of the European Conference on Computer Vision
  (ECCV)}, pages 786--803, 2018.

\bibitem{gurumurthy2017deligan}
Swaminathan Gurumurthy, Ravi Kiran~Sarvadevabhatla, and R Venkatesh~Babu.
\newblock Deligan: Generative adversarial networks for diverse and limited
  data.
\newblock In {\em Proceedings of the IEEE conference on computer vision and
  pattern recognition}, pages 166--174, 2017.

\bibitem{harvey2018recurrent}
Felix~G Harvey, Julien Roy, David Kanaa, and Christopher Pal.
\newblock Recurrent semi-supervised classification and constrained adversarial
  generation with motion capture data.
\newblock {\em Image and Vision Computing}, 78:42--52, 2018.

\bibitem{henter2020moglow}
Gustav~Eje Henter, Simon Alexanderson, and Jonas Beskow.
\newblock Moglow: Probabilistic and controllable motion synthesis using
  normalising flows.
\newblock {\em ACM Transactions on Graphics (TOG)}, 39(6):1--14, 2020.

\bibitem{hernandez2019human}
Alejandro Hernandez, Jurgen Gall, and Francesc Moreno-Noguer.
\newblock Human motion prediction via spatio-temporal inpainting.
\newblock In {\em Proceedings of the IEEE/CVF International Conference on
  Computer Vision}, pages 7134--7143, 2019.

\bibitem{holden2017phase}
Daniel Holden, Taku Komura, and Jun Saito.
\newblock Phase-functioned neural networks for character control.
\newblock {\em ACM Transactions on Graphics (TOG)}, 36(4):1--13, 2017.

\bibitem{holden2016deep}
Daniel Holden, Jun Saito, and Taku Komura.
\newblock A deep learning framework for character motion synthesis and editing.
\newblock {\em ACM Transactions on Graphics (TOG)}, 35(4):1--11, 2016.

\bibitem{holden2015learning}
Daniel Holden, Jun Saito, Taku Komura, and Thomas Joyce.
\newblock Learning motion manifolds with convolutional autoencoders.
\newblock In {\em SIGGRAPH Asia 2015 Technical Briefs}, pages 1--4. 2015.

\bibitem{ionescu2013human3}
Catalin Ionescu, Dragos Papava, Vlad Olaru, and Cristian Sminchisescu.
\newblock Human3. 6m: Large scale datasets and predictive methods for 3d human
  sensing in natural environments.
\newblock {\em IEEE transactions on pattern analysis and machine intelligence},
  36(7):1325--1339, 2013.

\bibitem{iskakov2019learnable}
Karim Iskakov, Egor Burkov, Victor Lempitsky, and Yury Malkov.
\newblock Learnable triangulation of human pose.
\newblock In {\em Proceedings of the IEEE/CVF International Conference on
  Computer Vision}, pages 7718--7727, 2019.

\bibitem{jang2020constructing}
Deok-Kyeong Jang and Sung-Hee Lee.
\newblock Constructing human motion manifold with sequential networks.
\newblock In {\em Computer Graphics Forum}, volume~39, pages 314--324. Wiley
  Online Library, 2020.

\bibitem{johansson1973visual}
Gunnar Johansson.
\newblock Visual perception of biological motion and a model for its analysis.
\newblock {\em Perception \& psychophysics}, 14(2):201--211, 1973.

\bibitem{kanazawa2019learning}
Angjoo Kanazawa, Jason~Y Zhang, Panna Felsen, and Jitendra Malik.
\newblock Learning 3d human dynamics from video.
\newblock In {\em Proceedings of the IEEE/CVF Conference on Computer Vision and
  Pattern Recognition}, pages 5614--5623, 2019.

\bibitem{karras2019style}
Tero Karras, Samuli Laine, and Timo Aila.
\newblock A style-based generator architecture for generative adversarial
  networks.
\newblock In {\em Proceedings of the IEEE/CVF Conference on Computer Vision and
  Pattern Recognition}, pages 4401--4410, 2019.

\bibitem{kaufmann2020convolutional}
Manuel Kaufmann, Emre Aksan, Jie Song, Fabrizio Pece, Remo Ziegler, and Otmar
  Hilliges.
\newblock Convolutional autoencoders for human motion infilling.
\newblock {\em arXiv preprint arXiv:2010.11531}, 2020.

\bibitem{kingma2014adam}
Diederik~P. Kingma and Jimmy Ba.
\newblock Adam: {A} method for stochastic optimization.
\newblock In Yoshua Bengio and Yann LeCun, editors, {\em 3rd International
  Conference on Learning Representations, {ICLR} 2015, San Diego, CA, USA, May
  7-9, 2015, Conference Track Proceedings}, 2015.

\bibitem{kingma2013auto}
Diederik~P Kingma and Max Welling.
\newblock Auto-encoding variational bayes.
\newblock {\em arXiv preprint arXiv:1312.6114}, 2013.

\bibitem{kundu2019bihmp}
Jogendra~Nath Kundu, Maharshi Gor, and R~Venkatesh Babu.
\newblock Bihmp-gan: Bidirectional 3d human motion prediction gan.
\newblock In {\em Proceedings of the AAAI conference on artificial
  intelligence}, volume~33, pages 8553--8560, 2019.

\bibitem{lebailly2020motion}
Tim Lebailly, Sena Kiciroglu, Mathieu Salzmann, Pascal Fua, and Wei Wang.
\newblock Motion prediction using temporal inception module.
\newblock In {\em Proceedings of the Asian Conference on Computer Vision},
  2020.

\bibitem{li2018convolutional}
Chen Li, Zhen Zhang, Wee~Sun Lee, and Gim~Hee Lee.
\newblock Convolutional sequence to sequence model for human dynamics.
\newblock In {\em Proceedings of the IEEE Conference on Computer Vision and
  Pattern Recognition}, pages 5226--5234, 2018.

\bibitem{li2020dynamic}
Maosen Li, Siheng Chen, Yangheng Zhao, Ya Zhang, Yanfeng Wang, and Qi Tian.
\newblock Dynamic multiscale graph neural networks for 3d skeleton based human
  motion prediction.
\newblock In {\em Proceedings of the IEEE/CVF Conference on Computer Vision and
  Pattern Recognition}, pages 214--223, 2020.

\bibitem{li2017auto}
Zimo Li, Yi Zhou, Shuangjiu Xiao, Chong He, Zeng Huang, and Hao Li.
\newblock Auto-conditioned recurrent networks for extended complex human motion
  synthesis.
\newblock {\em arXiv preprint arXiv:1707.05363}, 2017.

\bibitem{lin2018human}
Xiao Lin and Mohamed~R Amer.
\newblock Human motion modeling using dvgans.
\newblock {\em arXiv preprint arXiv:1804.10652}, 2018.

\bibitem{liu2018image}
Guilin Liu, Fitsum~A Reda, Kevin~J Shih, Ting-Chun Wang, Andrew Tao, and Bryan
  Catanzaro.
\newblock Image inpainting for irregular holes using partial convolutions.
\newblock In {\em Proceedings of the European Conference on Computer Vision
  (ECCV)}, pages 85--100, 2018.

\bibitem{mao2020history}
Wei Mao, Miaomiao Liu, and Mathieu Salzmann.
\newblock History repeats itself: Human motion prediction via motion attention.
\newblock In {\em European Conference on Computer Vision}, pages 474--489.
  Springer, 2020.

\bibitem{mao2019learning}
Wei Mao, Miaomiao Liu, Mathieu Salzmann, and Hongdong Li.
\newblock Learning trajectory dependencies for human motion prediction.
\newblock In {\em Proceedings of the IEEE/CVF International Conference on
  Computer Vision}, pages 9489--9497, 2019.

\bibitem{martinez2017human}
Julieta Martinez, Michael~J Black, and Javier Romero.
\newblock On human motion prediction using recurrent neural networks.
\newblock In {\em Proceedings of the IEEE Conference on Computer Vision and
  Pattern Recognition}, pages 2891--2900, 2017.

\bibitem{mildenhall2020nerf}
Ben Mildenhall, Pratul~P Srinivasan, Matthew Tancik, Jonathan~T Barron, Ravi
  Ramamoorthi, and Ren Ng.
\newblock Nerf: Representing scenes as neural radiance fields for view
  synthesis.
\newblock In {\em European conference on computer vision}, pages 405--421.
  Springer, 2020.

\bibitem{nazeri2019edgeconnect}
Kamyar Nazeri, Eric Ng, Tony Joseph, Faisal~Z Qureshi, and Mehran Ebrahimi.
\newblock Edgeconnect: Generative image inpainting with adversarial edge
  learning.
\newblock {\em arXiv preprint arXiv:1901.00212}, 2019.

\bibitem{pavlakos2019expressive}
Georgios Pavlakos, Vasileios Choutas, Nima Ghorbani, Timo Bolkart, Ahmed~AA
  Osman, Dimitrios Tzionas, and Michael~J Black.
\newblock Expressive body capture: 3d hands, face, and body from a single
  image.
\newblock In {\em Proceedings of the IEEE/CVF Conference on Computer Vision and
  Pattern Recognition}, pages 10975--10985, 2019.

\bibitem{pavllo2019modeling}
Dario Pavllo, Christoph Feichtenhofer, Michael Auli, and David Grangier.
\newblock Modeling human motion with quaternion-based neural networks.
\newblock {\em International Journal of Computer Vision}, pages 1--18, 2019.

\bibitem{pavllo20193d}
Dario Pavllo, Christoph Feichtenhofer, David Grangier, and Michael Auli.
\newblock 3d human pose estimation in video with temporal convolutions and
  semi-supervised training.
\newblock In {\em Proceedings of the IEEE/CVF Conference on Computer Vision and
  Pattern Recognition}, pages 7753--7762, 2019.

\bibitem{pavllo2018quaternet}
Dario Pavllo, David Grangier, and Michael Auli.
\newblock Quaternet: A quaternion-based recurrent model for human motion.
\newblock {\em arXiv preprint arXiv:1805.06485}, 2018.

\bibitem{rezende2015variational}
Danilo Rezende and Shakir Mohamed.
\newblock Variational inference with normalizing flows.
\newblock In {\em International Conference on Machine Learning}, pages
  1530--1538. PMLR, 2015.

\bibitem{rose1998verbs}
Charles Rose, Michael~F Cohen, and Bobby Bodenheimer.
\newblock Verbs and adverbs: Multidimensional motion interpolation.
\newblock {\em IEEE Computer Graphics and Applications}, 18(5):32--40, 1998.

\bibitem{rose2001artist}
Charles~F Rose~III, Peter-Pike~J Sloan, and Michael~F Cohen.
\newblock Artist-directed inverse-kinematics using radial basis function
  interpolation.
\newblock In {\em Computer Graphics Forum}, volume~20, pages 239--250. Wiley
  Online Library, 2001.

\bibitem{starke2019neural}
Sebastian Starke, He Zhang, Taku Komura, and Jun Saito.
\newblock Neural state machine for character-scene interactions.
\newblock {\em ACM Trans. Graph.}, 38(6):209--1, 2019.

\bibitem{su2021nerf}
Shih-Yang Su, Frank Yu, Michael Zollhoefer, and Helge Rhodin.
\newblock A-nerf: Surface-free human 3d pose refinement via neural rendering.
\newblock {\em arXiv preprint arXiv:2102.06199}, 2021.

\bibitem{szegedy2015going}
Christian Szegedy, Wei Liu, Yangqing Jia, Pierre Sermanet, Scott Reed, Dragomir
  Anguelov, Dumitru Erhan, Vincent Vanhoucke, and Andrew Rabinovich.
\newblock Going deeper with convolutions.
\newblock In {\em Proceedings of the IEEE conference on computer vision and
  pattern recognition}, pages 1--9, 2015.

\bibitem{tang2018long}
Yongyi Tang, Lin Ma, Wei Liu, and Weishi Zheng.
\newblock Long-term human motion prediction by modeling motion context and
  enhancing motion dynamic.
\newblock {\em arXiv preprint arXiv:1805.02513}, 2018.

\bibitem{taylor2007modeling}
Graham~W Taylor, Geoffrey~E Hinton, and Sam~T Roweis.
\newblock Modeling human motion using binary latent variables.
\newblock In {\em Advances in neural information processing systems}, pages
  1345--1352. Citeseer, 2007.

\bibitem{vaswani2017attention}
Ashish Vaswani, Noam Shazeer, Niki Parmar, Jakob Uszkoreit, Llion Jones,
  Aidan~N Gomez, Lukasz Kaiser, and Illia Polosukhin.
\newblock Attention is all you need.
\newblock {\em arXiv preprint arXiv:1706.03762}, 2017.

\bibitem{walker2017pose}
Jacob Walker, Kenneth Marino, Abhinav Gupta, and Martial Hebert.
\newblock The pose knows: Video forecasting by generating pose futures.
\newblock In {\em Proceedings of the IEEE international conference on computer
  vision}, pages 3332--3341, 2017.

\bibitem{wang2019t}
Tianming Wang and Xiaojun Wan.
\newblock T-cvae: Transformer-based conditioned variational autoencoder for
  story completion.
\newblock In {\em IJCAI}, pages 5233--5239, 2019.

\bibitem{wang2019combining}
Zhiyong Wang, Jinxiang Chai, and Shihong Xia.
\newblock Combining recurrent neural networks and adversarial training for
  human motion synthesis and control.
\newblock {\em IEEE transactions on visualization and computer graphics},
  27(1):14--28, 2019.

\bibitem{xiong2019foreground}
Wei Xiong, Jiahui Yu, Zhe Lin, Jimei Yang, Xin Lu, Connelly Barnes, and Jiebo
  Luo.
\newblock Foreground-aware image inpainting.
\newblock In {\em Proceedings of the IEEE/CVF Conference on Computer Vision and
  Pattern Recognition}, pages 5840--5848, 2019.

\bibitem{yan2018mt}
Xinchen Yan, Akash Rastogi, Ruben Villegas, Kalyan Sunkavalli, Eli Shechtman,
  Sunil Hadap, Ersin Yumer, and Honglak Lee.
\newblock Mt-vae: Learning motion transformations to generate multimodal human
  dynamics.
\newblock In {\em Proceedings of the European Conference on Computer Vision
  (ECCV)}, pages 265--281, 2018.

\bibitem{yu2018generative}
Jiahui Yu, Zhe Lin, Jimei Yang, Xiaohui Shen, Xin Lu, and Thomas~S Huang.
\newblock Generative image inpainting with contextual attention.
\newblock In {\em Proceedings of the IEEE conference on computer vision and
  pattern recognition}, pages 5505--5514, 2018.

\bibitem{yuan2019diverse}
Ye Yuan and Kris Kitani.
\newblock Diverse trajectory forecasting with determinantal point processes.
\newblock {\em arXiv preprint arXiv:1907.04967}, 2019.

\bibitem{yuan2020dlow}
Ye Yuan and Kris Kitani.
\newblock Dlow: Diversifying latent flows for diverse human motion prediction.
\newblock In {\em European Conference on Computer Vision}, pages 346--364.
  Springer, 2020.

\bibitem{zhang2018mode}
He Zhang, Sebastian Starke, Taku Komura, and Jun Saito.
\newblock Mode-adaptive neural networks for quadruped motion control.
\newblock {\em ACM Transactions on Graphics (TOG)}, 37(4):1--11, 2018.

\bibitem{zhang2019predicting}
Jason~Y Zhang, Panna Felsen, Angjoo Kanazawa, and Jitendra Malik.
\newblock Predicting 3d human dynamics from video.
\newblock In {\em Proceedings of the IEEE/CVF International Conference on
  Computer Vision}, pages 7114--7123, 2019.

\bibitem{zhang2021learning}
Siwei Zhang, Yan Zhang, Federica Bogo, Marc Pollefeys, and Siyu Tang.
\newblock Learning motion priors for 4d human body capture in 3d scenes.
\newblock In {\em Proceedings of the IEEE/CVF International Conference on
  Computer Vision}, pages 11343--11353, 2021.

\bibitem{zhang2020we}
Yan Zhang, Michael~J Black, and Siyu Tang.
\newblock We are more than our joints: Predicting how 3d bodies move.
\newblock {\em arXiv preprint arXiv:2012.00619}, 2020.

\bibitem{zheng2019pluralistic}
Chuanxia Zheng, Tat-Jen Cham, and Jianfei Cai.
\newblock Pluralistic image completion.
\newblock In {\em Proceedings of the IEEE/CVF Conference on Computer Vision and
  Pattern Recognition}, pages 1438--1447, 2019.

\end{thebibliography}
}

\newpage

\appendix
\setcounter{page}{1}

\twocolumn[
\centering
\Large
\textbf{\textsc{TrajeVAE}: Controllable Human Motion Generation from Trajectories} \\
\vspace{0.5em}Supplementary Material \\
\vspace{1.0em}
] %
\appendix

\section{Adapting MoGlow}
As mentioned in the main text, MoGlow \cite{henter2020moglow} conditions predicted poses on a \textit{control signal}. This signal represents relative and rotational velocities on the ground plane. Moreover, the authors use the exponential map representation but at the same time claim that MoGlow can be used for any other well-known skeleton representation. We identify the following changes to the original implementation that enabled us to use MoGlow in our framework:
\begin{enumerate}
    \item We change the exponential map representation to the 3D coordinates of $J$ joints.
    \item We replace the control signal represented as velocities into trajectories defined as poses with some of the joints set to 0. This increases the input signal's dimensionality from $3T$ to $3JT$ for $T$ time steps. However, it has a negligible effect on the performance.
    \item We also removed regularization techniques such as gradient norm clipping and gradient value clipping and disabled data normalization. These techniques deteriorate the learning, and the network does not converge in our scenario.
    \item For consistency with other methods, we use the Adam optimizer \cite{kingma2014adam} with the same learning rate regime.
\end{enumerate}
We left the rest of the implementation unchanged. 

\section{Implementation details}
\paragraph{\trajevae{}} MLPs applied in the input and in the CVAE's bottleneck output latent codes of size $256$. Therefore, vectors $\hat{\mathbf{H}}$ and $\bar{\mathbf{H}}$ processed by self-attention layers have a dimensionality $512$. The initial layer in the decoder $\mathcal{D}$ processes $\{[\mathbf{w}_t; \mathbf{w}_0]\}$ and is defined as a function: $f: \mathbb{R}^{768} \rightarrow \mathbb{R}^{512}$. The final layers outputs vectors of size $3J$.

All MLPs responsible for encoding poses and trajectories mentioned in the main text consists of the following structure: \code{Linear} $\rightarrow$ \code{Layer Normalization} $\rightarrow$ \code{Leaky ReLU($\alpha$=0.1)} $\rightarrow$ \code{Linear} $\rightarrow$ \code{Layer Normalization} $\rightarrow$ \code{Leaky ReLU($\alpha$=0.1)}, where $\alpha$ is a scale of the negative slope of the function. The initial MLP in the decoder $\mathcal{D}$  has the structure \code{Linear} $\rightarrow$ \code{Layer Normalization} $\rightarrow$ \code{Leaky ReLU($\alpha$=0.1)} $\rightarrow$ \code{Linear}.

The CVAE baseline operate with the same dimensionalities as \trajevae{}. We implement them in the same way as defined in \cite{yuan2020dlow}. The model uses GRU network to encode the temporal data. The recurrent decoder also receives a coordinate of the trajectory $\mathbf{y}_t$ in the time step $t$. 

We additionally apply dropout $=0.1$ to self-attention layers as described in \cite{vaswani2017attention}.

\section{Same pose, different trajectories}
We perform an additional experiment that confirms the generality of our approach.  We show results for a scenario when we use different trajectories for the same initial pose. As expected, the generated poses follow different trajectories even though such combinations do not occur in the dataset.

\paragraph{Preparing the data}
To maintain plausibility that a particular initial pose is physically capable of following a conditioning trajectory, we pair each initial pose $\mathbf{x}_0$ in the dataset with all trajectories where the distance between $\mathbf{x}_0$ and coordinates of the trajectory in a time step $t= 0$ is below $\epsilon_0 = 0.01$. Since obtaining the ground truth sequence $\mathbf{X}$ in such a case is not possible, we assume that the sequence $\mathbf{X}$ corresponding to a given trajectory is a sufficient approximation of the expected sequence. We evaluate \trajevae{} as previously using APD, ADE, FDE. We omit MMADE and MMFDE for its exponential computational complexity that this scenario creates.

\paragraph{Results} We present results in Tab.~\ref{tab:different-scenario}. Even though these trajectories do not come from the same sequence as the initial poses, \trajevae{} generates a sequence that follows the trajectory. The decrease in accuracy (ADE) between $k = 2$ and $k=3$ is caused by adding a trajectory that corresponds to the right hand, while $k < 2$ we add only trajectories regarding feet. While feet commonly behave similarly throughout the animation, hands have a significantly different motion from other joints.

The value $k=0$ corresponds to no trajectories, and therefore we omit it in the Tab.~\ref{tab:different-scenario}. Refer to supplementary files to find animations generated for initial poses with different trajectories.

\begin{table}[!htbp]
    \centering
    \small
    \begin{tabular}{cccc}
    \toprule
         $k$ & APD &    ADE &    FDE \\
        \midrule
         1 &     5.373 &  0.370 &  0.476 \\
         2 &     5.400 &  0.362 &  0.472 \\
         3 &     5.096 &  0.375 &  0.491 \\
         4 &     4.175 &  0.332 &  0.433 \\
    \bottomrule
    \end{tabular}
    \caption{
    \textbf{Quantitative results} for the Human3.6M dataset when $K=50$ samples are generated for the scenario where we use different trajectories from the whole dataset for the same initial pose. We assume different situations that \mbox{$k=\{1, 2, 3, 4\}$} trajectories are provided.}
    \label{tab:different-scenario}
\end{table}

\section{Extended ablation study}
In experiments, we provide results for an ablation study when only a trajectory for the right foot is provided. We additionally show in Tab.~\ref{tab:additional-ablation} results when we input no trajectories, or progressively add trajectories of the right foot, left foot, right hand, and left hand. The extended results show that our design decisions consistently affect scenarios when we vary the number of the input trajectories.

\begin{table}[htbp!]
    \centering
    \notsotiny
\begin{tabular}{lcccccc}
    \toprule
         \multirow{2}{*}{Method} & \multirow{2}{*}{$k$} & APD  & ADE  & FDE  & MMADE  & MMFDE  \\ 
         & & $\uparrow$ & $\downarrow$ & $\downarrow$& $\downarrow$& $\downarrow$\\
        \midrule
                  Base &  \multirow{5}{*}{0} &           1.237 &           0.525 &           0.749 &           0.634 &           0.801 \\
     + Learnable prior &   &           1.860 &  \textbf{0.502} &  \textbf{0.694} &  \textbf{0.611} &  \textbf{0.740} \\
                 + DCT &   &  \textbf{9.483} &           0.560 &           0.742 &           0.634 &           0.762 \\
 + Masked future poses &   &           8.936 &           0.539 &           0.724 &           0.616 &           0.748 \\
        \midrule
                  Base &  \multirow{5}{*}{1} &           1.220 &           0.481 &           0.695 &           0.640 &           0.811 \\
     + Learnable prior &   &           1.749 &  \textbf{0.448} &  \textbf{0.622} &           0.618 &           0.753 \\
                 + DCT &   &  \textbf{7.014} &           0.487 &           0.637 &           0.602 &           0.703 \\
 + Masked future poses &   &           6.803 &           0.472 &           0.623 &  \textbf{0.594} &  \textbf{0.695} \\
        \midrule
                  Base &  \multirow{5}{*}{2} &           1.217 &           0.466 &           0.672 &           0.642 &           0.812 \\
     + Learnable prior &   &           1.719 &  \textbf{0.418} &  \textbf{0.577} &           0.616 &           0.750 \\
                 + DCT &   &  \textbf{6.561} &           0.465 &           0.604 &           0.594 &           0.688 \\
 + Masked future poses &   &           6.286 &           0.448 &           0.587 &  \textbf{0.587} &  \textbf{0.681} \\
        \midrule
                  Base &  \multirow{5}{*}{3} &           1.248 &           0.376 &           0.549 &           0.639 &           0.802 \\
     + Learnable prior &   &           1.725 &  \textbf{0.327} &  \textbf{0.449} &           0.632 &           0.767 \\
                 + DCT &   &  \textbf{5.367} &           0.389 &           0.503 &           0.590 &           0.681 \\
 + Masked future poses &   &           4.861 &           0.361 &           0.474 &  \textbf{0.585} &  \textbf{0.674} \\
        \midrule
                  Base &  \multirow{5}{*}{4} &           1.261 &           0.339 &           0.498 &           0.641 &           0.805 \\
     + Learnable prior &   &           1.720 &  \textbf{0.281} &  \textbf{0.385} &           0.646 &           0.783 \\
                 + DCT &   &  \textbf{4.524} &           0.338 &           0.443 &           0.595 &           0.692 \\
 + Masked future poses &   &           3.857 &           0.312 &           0.412 &  \textbf{0.594} &  \textbf{0.689} \\
    \bottomrule
    \end{tabular}
    \caption{
        Influence of design decisions on obtained results for \mbox{$k=\{0, 1,2 ,3, 4\}$} trajectories. These trajectories refer to scenarios when we use no trajectories and then add progressively trajectories for the right foot, left foot, right hand, and left hand. The best results are in bold. 
    }
    \label{tab:additional-ablation}
\end{table}

\end{document}